\pdfoutput=1

\documentclass[11pt]{article}

\usepackage[final]{acl}

\usepackage{times}
\usepackage{latexsym}

\usepackage[T1]{fontenc}

\usepackage[utf8]{inputenc}

\usepackage{microtype}

\usepackage{inconsolata}

\usepackage{graphicx}

\usepackage{hyperref}
\usepackage{url}
\usepackage{wrapfig}
\usepackage{tcolorbox}
\usepackage{xcolor}
\usepackage{bm}
\usepackage{amssymb}
\usepackage{amsmath}
\usepackage{amsfonts}
\usepackage{graphicx}
\usepackage{subcaption}
\usepackage{dblfloatfix}
\usepackage{float}
\usepackage{chngcntr}

\title{Features that Make a Difference: \\ Leveraging Gradients for Improved Dictionary Learning}

\author{Jeffrey Olmo \and Jared Wilson \and Max Forsey \and Bryce Hepner  \\
        \and {\bf Thomas Vin Howe} \and {\bf David Wingate} \\
        {\bf Brigham Young University}\\}

\begin{document}
\maketitle
\begin{abstract}
Sparse Autoencoders (SAEs) are a promising approach for extracting neural network features by learning a sparse and overcomplete decomposition of the network's internal activations. 
However, SAEs are traditionally trained considering only activation values and not the effect those activations have on downstream computations. 
This limits the information available to learn features, and biases the autoencoder towards neglecting features which are represented with small activation values but strongly influence model outputs.
To address this, we introduce Gradient SAEs (g-SAEs), which modify the $k$-sparse autoencoder architecture by augmenting the TopK activation function to rely on the gradients of the input activation when selecting the $k$ elements.  
For a given sparsity level, g-SAEs produce reconstructions that are more faithful to original network performance when propagated through the network.
Additionally, we find evidence that g-SAEs learn latents that are on average more effective at steering models in arbitrary contexts.
By considering the downstream effects of activations, our approach leverages the dual nature of neural network features as both \textit{representations}, retrospectively, and \textit{actions}, prospectively. While previous methods have approached the problem of feature discovery primarily focused on the former aspect, g-SAEs represent a step towards accounting for the latter as well. 
\end{abstract}

\section{Introduction}

Sparse Autoencoders (SAEs) have emerged as a promising method for interpreting neural networks, aiming to recover a model's \textit{features} via dictionary learning \citep{Bricken2023Towards,cunningham_sparse_2023, templeton2024scaling}.
While there is no universally accepted definition of features, they are generally understood to be the atomic units of language models' computation, possessing the quality of monosemanticity as both representations and causal intermediates. 
The success of SAEs in identifying directions in language model activation space that are causally relevant and interpretable provides evidence that a significant fraction of the internal activations of language models are sparse, linear combinations of vectors which are each oriented in a direction corresponding to a feature of the model. \citep{park2024linearrepresentationhypothesisgeometry, park_geometry_2024}. The elements of the SAE-learned dictionary corresponding to these directions we call \textit{latents}. 

However, SAEs likely have room for improvement. SAEs  are trained to encode input activations such that the reconstruction loss over many tokens from a training corpus is minimized.
This has raised some concerns that SAEs may not learn latents corresponding to the features of a model, but rather the features of a dataset \citep{e2e_sae_braun, dooms_tokenized_2024}.
That is, SAEs might give the illusion of interpretability by learning latents  based on frequent and distinct concepts in the \textit{training corpus}, rather than learning latents because they correspond to features that play an important and distinct role in the model's decision-making process.
We would like a better guarantee that our interpretability tools discover the features of a model, rather than primarily reflecting the training data.

Additionally, if one is interested in interpretability as a means to exert more fine-grained control over the behavior of models, it would be desirable to have interpretability tools that are biased towards uncovering the features that are most responsible for a model's output. \citet{templeton2024scaling} speculate that current Large Language Models (LLMs) may represent orders of magnitude more features than the size of the dictionaries of the largest SAEs trained to date.
The limited capacity of SAEs incentivizes them to neglect features that strongly affect model outputs if overlooking them usually results in small reconstruction errors \citep{e2e_sae_braun}.

\begin{figure*}
    \centering
    \includegraphics[width=0.80\linewidth]{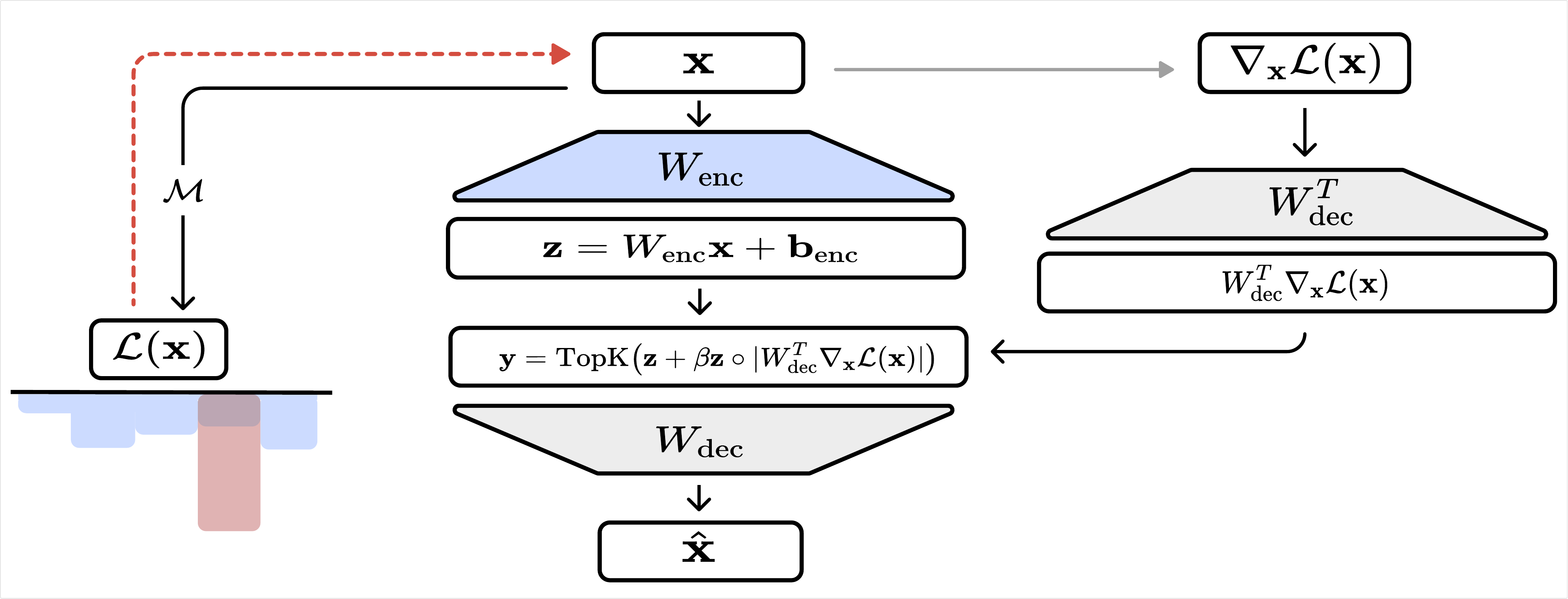}
    \caption{The setup of g-SAE training. Here $\mathcal{L}(\mathbf{x})$ is the function mapping the residual stream activation at the layer the SAE was trained on, to the predictive cross entropy loss it yields. The red dotted line denotes backpropagation.}
    \label{fig:g-SAE_setup}
\end{figure*}

These considerations motivate our approach, which aims to address the challenges above by connecting dictionary learning more closely with model \textit{outputs}. The resulting architecture we call g-SAEs. In this paper we show:

\begin{enumerate}
    \item \textbf{Distance in LLM activation space alone correlates poorly with difference in LLM output.}
    We show that there is only a weak correlation between the norm of a perturbation vector and it's effect on LLM downstream outputs, as measured by change in prediction cross-entropy loss. 
    In contrast to the norm alone, the first order approximation given by gradients of activations with respect to loss strongly predict changes in loss for perturbations of moderate size.
    This suggests that reconstruction loss minimization doesn't directly incentivize the SAE to learn reconstructions that minimize difference in effect on outputs. 

    \item \textbf{In SAE training, incorporating gradients produces improvements on various Pareto frontiers defined by popular SAE architectures.} 
    For a given expansion size, using a gradient-aware TopK activation function leads to improvements in the minimization of difference in model loss when reconstructed outputs are propagated downstream.
    It also leads to fewer permanently inactive units of the SAE hidden layer.   

    \item \textbf{g-SAE latents provide superior steering with no cost to interpretability.}
    Most promisingly, g-SAEs recover latents that exert a more \textit{influential} and \textit{targeted} effect on logits they are associated with when used as steering vectors, while remaining as interpretable as the latents of other SAEs.

\end{enumerate}

\section{Motivation and Background}

\subsection{SAE Preliminaries}

The core of an SAE is the following: an encoder matrix which projects the input activation into a higher dimensional latent space, where the elements represent a dictionary of learned features, and a decoder which reconstructs the input from this sparse vector.
Formally, for a model's activation vector $\mathbf{x} \in \mathbb{R}^d$,
\begin{gather*}
\mathbf{y} = \text{ReLU}(W_\text{enc}(\mathbf{x} - \mathbf{b}_\text{dec}) + \mathbf{b}_\text{enc})  \\
\hat{\mathbf{x}} = W_\text{dec}\mathbf{y} + \mathbf{b}_\text{dec} 
\end{gather*}
where $W_\text{enc}\mathbf{x} \in \mathbb{R}^h$, $h$ is the size of the dictionary, and 
$W_\text{enc} \in \mathbb{R}^{h \times d}$.
We will denote the sparse latent vector $W_\text{enc}\mathbf{x}$ as $\mathbf{y}$ and the reconstructed activation as $\hat{\mathbf{x}}$.
The elements of $\mathbf{y}$ are the SAE's latents.

Generally, SAEs are trained with a loss function defined by the minimization of the reconstruction error $\|\mathbf{x}-\hat{\mathbf{x}}\|_2^2$ and some mechanism for promoting sparsity.
To this end, \citet{Bricken2023Towards} use a term in the loss function to minimize the $L_1$ norm of $\mathbf{y}$.
This approach comes with some downsides, including the phenomenon of \textit{shrinkage}, that is, the unintended optimization of the elements of $\mathbf{y}$ to be minimized in addition to promoting the sparsity of $\mathbf{y}$.
Various alternative architectures have been proposed, which we detail briefly below. 

\subsection{Relevant Work}

\begin{figure}
    \centering
    \includegraphics[width=.95\linewidth]{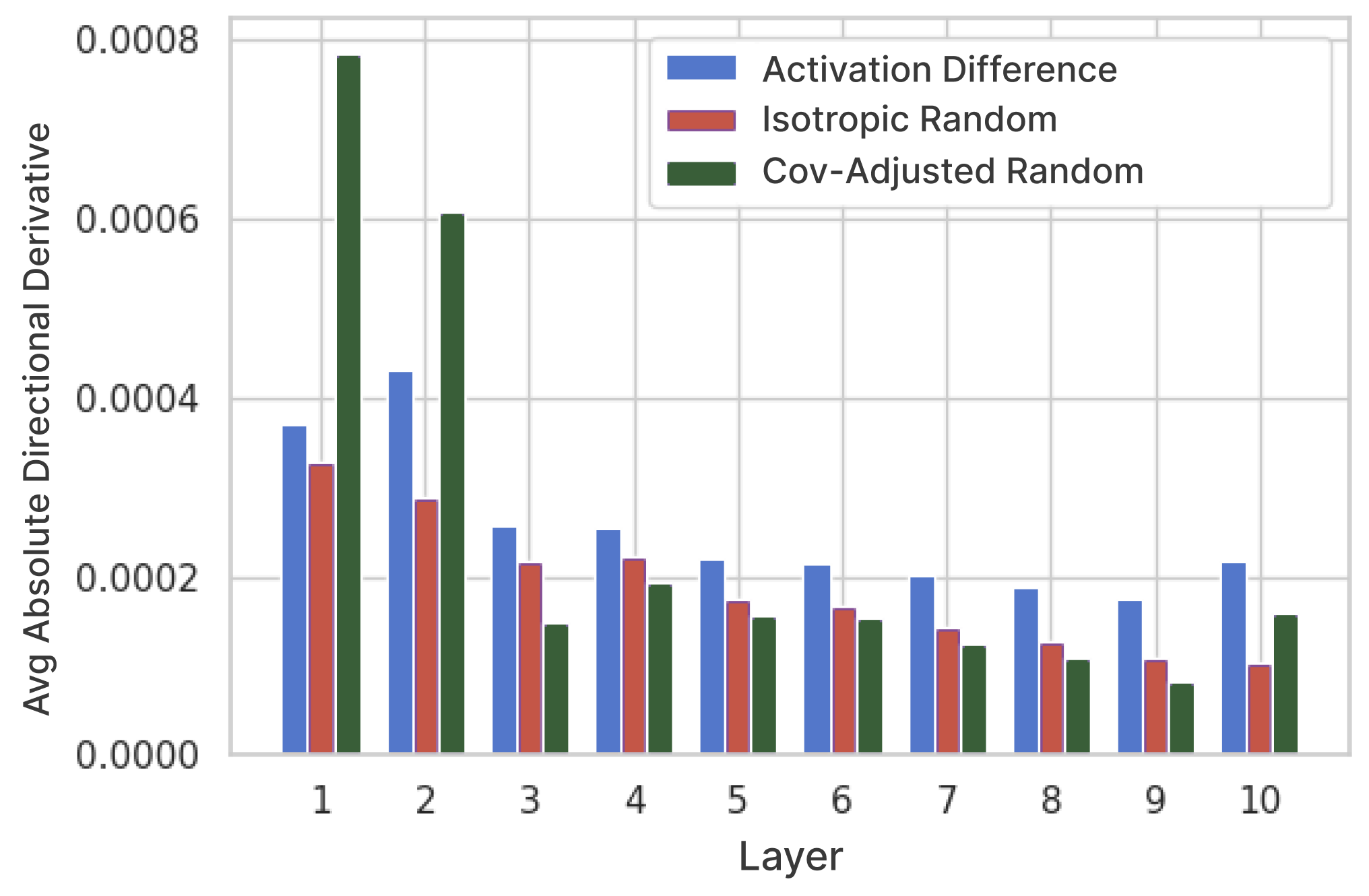}
    \caption{Comparison of directional derivatives in  GPT-2's residual stream. The plot shows the absolute value of the directional derivatives towards activation differences vs. random directions. Layer $n$ corresponds to the $n_\text{th}$ \texttt{resid\_post} hookpoint. In later layers, feature directions exhibit consistently higher derivatives with respect to loss.}
    \label{fig:projections}
\end{figure}

\citet{gao_scaling_2024} introduced the TopK autoencoders architecture which use a TopK activation function to serve the purpose of direct sparsity enforcement.
This activation function zeroes out all but the $k$ most active elements of the pre-activation $\mathbf{z}$.
Various other architectures have been proposed in addition to the TopK architecture to improve upon the basic sparse autoencoder setup, including Gated SAEs, JumpReLU SAEs, Tokenized SAEs and Switch SAEs \citet{rajamanoharan2024improvingdictionarylearninggated, rajamanoharan2024jumpingaheadimprovingreconstruction, dooms_tokenized_2024, mudide2024efficientdictionarylearningswitch}, \citep{rajamanoharan2024jumpingaheadimprovingreconstruction}. 
Recent research has also explored the challenge of identifying functionally significant features.
In particular, \citet{e2e_sae_braun} posited that reconstruction loss may not strongly incentivise an SAE to learn latents that explain a model's performance.
To address this limitation, they developed end-to-end (e2e) SAEs.
Rather than using MSE as a loss function, e2e SAEs optimize for explanation of downstream effect by minimizing the KL divergence between the output logits of the original activations and those of the SAE-reconstructed activations.
However, e2e SAE latents appear to be less influential compared to those derived from standard SAE architectures.
\citet{lee2024investigatingsensitivedirections} hypothesize that this is because e2e SAEs optimize their objective by exploiting the space outside the typical activation space.

\subsection{LLM outputs are particularly locally sensitive to perturbations of activations in feature directions}

\begin{figure*}
    \centering
    \begin{minipage}[t]{0.47\linewidth}
        \centering
        \includegraphics[width=\linewidth]{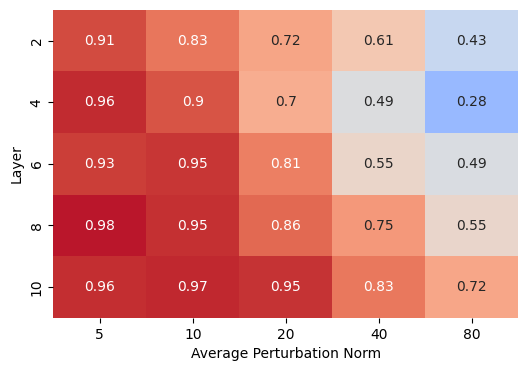}
    \end{minipage}
    \hspace{0.02\linewidth} 
    \begin{minipage}[t]{0.47\linewidth}
        \centering
        \includegraphics[width=\linewidth]{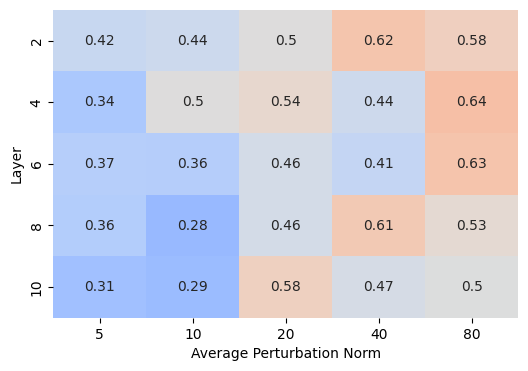}
    \end{minipage}
    \caption{\textbf{Left:} Spearman correlation coefficients between change in cross-entropy loss $\Delta \mathcal{L}(\mathbf{x})$ from isotropically perturbing an MLP output, and the first order approximation of that change ($|\nabla_\mathbf{x}{\mathcal{L}(\mathbf{x})} \cdot \delta{\mathbf{x}}|$). \textbf{Right:} Correlations between the norm of a perturbation and the resulting effect on loss. Perturbations for each column are drawn from a uniform distribution with the displayed mean and a standard deviation of one-half the mean. Data from \texttt{resid\_post} hookpoint of GPT-2 across diverse tokens.}
    \label{fig:combined}
\end{figure*}

The linear representation hypothesis \citep{park2024linearrepresentationhypothesisgeometry, park_geometry_2024, elhage2022superposition} posits that deep learning models represent features as linear directions in activation space. Assuming this is the case, then it follows that the difference between two real activation vectors corresponds to a linear combination of feature directions. Previous work has analyzed the sensitivity of GPT-2's \citep{radford2019language} output to perturbations in random directions vs. perturbations in the direction of the difference between two real activation vectors \citep{heimersheim2024activationplateaus, giglemiani2024evaluatingsyntheticactivationscomposed}.
This work suggests that LLM outputs are significantly more sensitive to finite perturbations in the direction of linear combinations of features than perturbations in random directions.

We tested whether this heightened sensitivity to linear combinations of feature directions implies that the directional derivative of activations in these directions is particularly large as well; for arbitrary residual stream activations $\mathbf{x}$ of GPT-2 within a sequence, we compute $\nabla_\mathbf{x}\mathcal{L}(\mathbf{x})$ where $\mathcal{L}$ is the function mapping $\mathbf{x}$ to the corresponding cross entropy loss it yields.
We then compare the average directional derivative in random directions and in the direction of arbitrary activation differences $\mathbf{x}_i - \mathbf{x}_j$. 
We use two different kinds of random directions; isotropically random directions\footnote{Here, isotropic means that the direction of the permutation is uniformly random across components in the standard basis, such that no component is favored.}, and covariance-adjusted random directions that mirror the covariance structure of real activations \citep{lee2024investigatingsensitivedirections}.

For layers 3 onward, where we expect sophisticated features to exist, the activation difference directions have significantly higher gradient products than both isotropic random and covariance-adjusted random directions, illustrated in Figure \ref{fig:projections}.
This confirms previous findings that the local linear approximation of output change provided by gradients are accurate on the relevant scales \citep{syed2023attributionpatchingoutperformsautomated, templeton2024scaling}.
This consistent pattern for later layers motivates the key hypothesis behind our method:

\textit{A model's output is particularly locally sensitive towards feature directions.}

\subsection{Gradient-weighted distance in activation space is a strong predictor of difference in output}
Standard SAEs use activation difference norms to learn their decompositions of model activations.
Are norms in activation space a reliable guide to degree of influence on outputs?

We find that in all MLP output layers of the LLMs we tested, there is generally a weak correlation between the norm of an isotropic random permutation to an MLP output, and the resulting change to prediction cross-entropy loss (Fig. \ref{fig:combined}).
In the context of SAEs, this low correlation means that ``not all reconstruction errors are equal.''
Frequently, small  perturbations will cause large downstream effects on loss, and conversely, large perturbations can be ineffectual. Indeed, we find that the effect of a given perturbation can vary by as much as 3 orders of magnitude depending on its direction. The standard SAE training setup makes no distinction between reconstruction errors if they have the same norm.
Additionally, conditional upon model outputs being especially sensitive in feature directions, the standard SAE training setup leaves valuable information on the table in the search for features.

On the other hand, the first order approximation of the perturbation which leverages the gradient has, in general, a stronger correlation with the resulting effect, especially at later layers and smaller perturbations.
Given that SAEs generally produce reconstruction errors with norms on the lower end of this range, we hypothesize that in the context of SAE training, gradients carry reliable information about the relative influence on model output of different directions in activation space, which we can leverage for better dictionary learning.

\begin{figure*}
    \centering
    \includegraphics[width=0.95\linewidth]{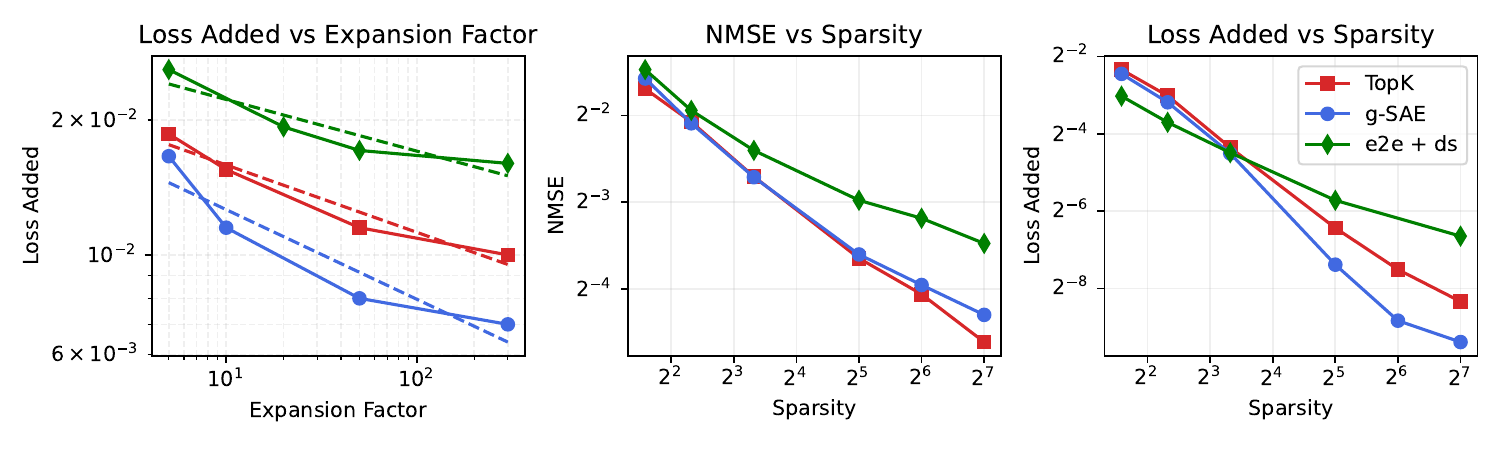}
    \caption{$\textbf{Left:}$ Number of latents vs Loss Added ($\mathcal{L}_{\text{added}}$) for $L_0$=32. $\textbf{Middle:}$ $L_0$ against NMSE holding the number of latents fixed to 15360. $\textbf{Right:}$ $L_0$ against Loss Added, holding number of latents fixed.
    All SAEs were trained on GPT-2 small with $\sim$14M tokens.}
    \label{fig:pareto_frontiers}
\end{figure*}

\section{g-SAEs}
A g-SAE is a standard SAE but with the following activation function: 

\begin{equation*}
\sigma(\mathbf{z})_i =
\begin{cases}
\mathbf{z}_i & \text{if } i \in K\\
0 & \text{otherwise}
\end{cases}
\end{equation*}
$$K = \text{TopK}\left(\mathbf{z} + \beta \mathbf{z} \circ \left|W_\text{dec}^T \cdot \nabla_\mathbf{x}\mathcal{L}(\mathbf{x})\right|\right)$$

\label{eq:attribution-topK}

where $\mathcal{L}(\mathbf{x})$ is the function mapping a residual stream activation to its associated predictive cross entropy loss, $W_\text{dec}$ is the decoder matrix, and $\beta$ is a tunable hyperparameter.
This activation function selects $k$ latents $\mathbf{z}_i$ based on two criteria: 1) the value of $\mathbf{z}_i$, and 2) $\mathbf{z}_i$ times the $i_\text{th}$ component of $W_\text{dec}^T \cdot \nabla_\mathbf{x}\mathcal{L}(\mathbf{x})$, which is the linear transformation of the gradient at $\mathbf{x}$ into the SAE latent space.
The latter term can be thought of as the attribution of $\mathbf{z}$ with respect to the model loss; it resembles the attribution vector derived by attribution patching \citep{nanda2024attributionpatching}\footnote{Key differences between attribution here and attribution in \citet{nanda2024attributionpatching} are that we effectively use a baseline value of 0 for the feature instead of a baseline value taken from the feature’s activity on a second prompt. Additionally, gradients of $\mathbf{z}$ are never taken, rather the gradient of x is projected into the SAE latent space with the decoder}.
The term $\mathbf{z} \circ \left|W_\text{dec}^T \cdot \nabla_\mathbf{x}\mathcal{L}(\mathbf{x})\right|$ represents the local linear approximation of the effect that turning latent $\mathbf{z}_i$ off has on the loss yielded by $\mathbf{x}$.

These dual criteria place productive constraints on both the encoder and decoder: when $\beta$ is high, the decoder is forced to work with only latents corresponding to directions in the input space with large effects as estimated by the local linear approximation, and therefore learn to reconstruct $\mathbf{x}$ by leveraging these directions.
Likewise, with high $\beta$, the encoder is only updated along rows corresponding to high-influence latents, (because in backpropagation, no gradients flow through the zeroed-out latents) and consequently, it must learn a decomposition of $\mathbf{x}$ along directions with high attributions with respect to the model's loss. 
Thus, while the SAE loss remains reconstruction error, the SAE is biased towards learning latents oriented towards high-influence directions.
A major advantage of our method is that this is accomplished without incorporating gradients with respect to $\mathbf{x}$ into the loss function of the SAE, which we find tends to be unstable and expensive.

The experiments in this section are all performed on layer 7 of GPT-2 small.\footnote{Throughout training, a variety of machines were used including one with 8 A100 GPUs training for two to three hours per model.}
 For g-SAEs, we use $\beta = 10^5$. Following \citet{dooms_tokenized_2024}, we use the added cross-entropy loss to measure the impact on the model prediction and normalized MSE to measure reconstruction:

\small
\begin{minipage}{0.05\textwidth}
    \begin{align*}
        \mathcal{L}_{\text{added}}(\mathbf{x}) = \frac{\mathcal{L}(\mathbf{\hat{x}}) - \mathcal{L}(\mathbf{x})}{\mathcal{L}(\mathbf{x})}, 
    \end{align*}
\end{minipage}%
\begin{minipage}{0.05\textwidth}
    \begin{align*}
        \text{  NMSE}(\mathbf{x}) = \frac{\|\mathbf{x} - \mathbf{\hat{x}}\|_2}{\|\mathbf{x}\|_2}
    \end{align*}
\end{minipage}
\normalsize

\section{Results}
\subsection{Capacity}

\begin{figure*}[t]
    \centering
    \includegraphics[width=\linewidth]{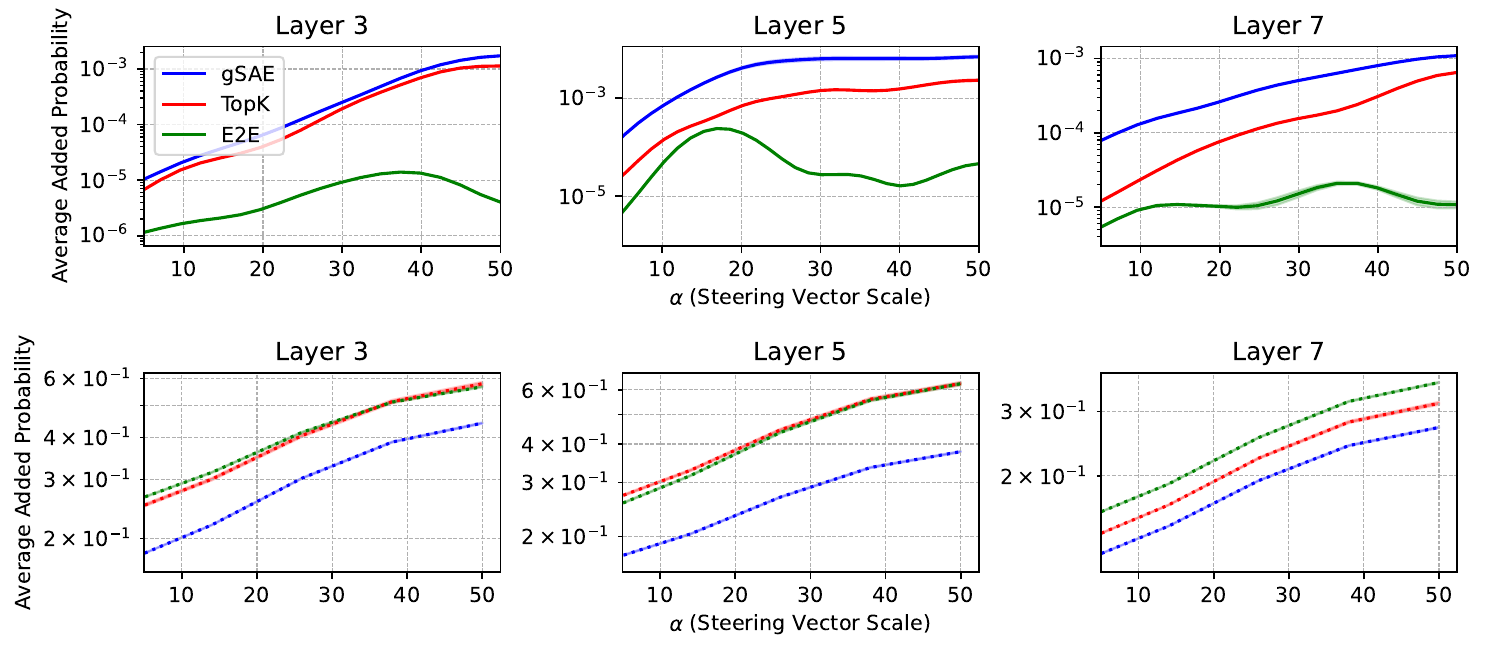}
    \caption{\textbf{Top:} The average effect of applying a steering vector in the direction of a latent $\mathbf{y}_i$ on the logits $\mathbf{y}_i$ points towards. \textit{Higher is better.} \textbf{Bottom:} The average total probability added to all other logits when applying a steering vector in the direction of $\mathbf{y}_i$. \textit{Lower is better.} Data from a random samples of latents from SAEs with $L_0 = 32$, with $\alpha$ incremented in steps of 5/3 above, and 10 below. Gaussian smoothing applied for better visibility.}
    \label{fig:control_and_specificity}
\end{figure*}

g-SAEs perform comparably on the NMSE-sparsity frontier, and add slightly less loss for a given sparsity, particularly at denser sparsities.
When sparsity is held constant and expansion size varies, g-SAEs display a steeper curve than TopK, indicating increased efficiency in utilizing additional capacity.
This suggests that g-SAEs leverage additional latents more productively than pure TopK SAEs. 

Related to this is the fact that g-SAEs have fewer inactive, or ``dead'' latents than other architectures, for a given sparsity.

We find that when $\beta$ is too large, performance degrades in terms of NMSE and loss added, suggesting that both terms in the activation function $\sigma$ are necessary. 

\begin{table}[h]
\centering
\begin{tabular}{lccc}
\hline
\textbf{Sparsity} & \textbf{g-SAE} & \textbf{TopK} & \textbf{E2E} \\
\hline
3 & 0.95 & 0.98 & 0.95 \\
5 & 0.88 & 0.92 & 0.87 \\
10 & 0.71 & 0.81 & 0.68 \\
32 & 0.34 & 0.43 & 0.29 \\
64 & 0.19 & 0.26 & 0.16 \\
\hline
\end{tabular}
\caption{A comparison of dead latents against $L_0$ sparsity. Here a latent is considered dead if it doesn't activate for 5 consecutive batches.}
\label{fig:dead_latents}
\end{table}

\subsection{Steering}

A core motivation behind g-SAEs is the desire to learn latents which have an interpretable and strong causal connection with the model's outputs.
Such a connection would be useful in providing more control over model behavior through the use of latent directions as steering vectors \citep{li2024inferencetimeinterventionelicitingtruthful, turner2024activationadditionsteeringlanguage, marks2024geometrytruthemergentlinear, panickssery2024steeringllama2contrastive, conmy2024activationsteering, templeton2024scaling}.
Additionally, evidence of a causal relationship between latent activations and model outputs would suggest that latents are more closely tracking model features, if we assume that features have a dual nature as both representations and actions. 

To investigate this, for a given alive latent $\mathbf{y}_i$, we collected the set $L$ of the $n$ logits that $\mathbf{y}_i$ is most associated with: $L$ is computed by projecting $\mathbf{y}_i$'s direction into the vocabulary space via the unembedding matrix $W_U$, yielding a logit vector $W_U^\top \mathbf{y}_i$, and then selecting the $n$ logits corresponding to the largest values in $W_U^\top \mathbf{y}_i$, which represent the vocabulary tokens that $\mathbf{y}_i$ points towards most.
Then, we compute the following:
$$\sum_{j \in L} \left(\mathcal{M}\left(\mathbf{x} + \alpha \frac{W_\text{dec}^i}{\|W_\text{dec}^i\|}\right)_j - \mathcal{M}(\mathbf{x})_j\right)$$

Where: \\
- $L$ is the set of associated logits for latent $\mathbf{y}_i$, \\
- $\mathcal{M}$ is the function mapping the current layer to the output probability vector \\
- $\mathbf{x}$ is the original activation, \\
- $\alpha$ is the scaling factor, \\
- $W_\text{dec}^i$ is the $i_\text{th}$ decoder column \\

Conceptually, this equation represents the degree of latent-specific influence a steering vector exerts over the output.
We see that on average, g-SAE latents more strongly cause increases to their corresponding logits when used as steering directions.
This increase is substantial: the added probability of a corresponding logit is on average over 5x greater in g-SAEs than in TopK SAEs (see Fig. \ref{fig:control_and_specificity}).
Additionally, we tested the specificity of the effect of adding the latent-derived steering vectors by measuring the added probability to all \textit{except} the $n$ most associated logits.
Here we see that g-SAEs again display superior performance; they increase the likelihood of arbitrary logits less than other SAEs' steering vectors. 

Focusing on just $\alpha = 50$ for layer 7, we plot the kernel density estimation plots for the added probability to the associated logits for 1000 latents from each SAE, over 10 arbitrary input sequences each.
We see in the top plot of Fig.~\ref{fig:KDEs} that the right tails of the TopK and g-SAE distributions look quite similar, and that the difference lies in the fact that g-SAEs have fewer latents that fail to affect their associated logits, shifting the mean of the distribution to the right. This suggests that g-SAEs learn more latents which are on the \textit{natural} upper end of the influence distribution, rather than learning latents which have \textit{unnaturally} high influence on outputs (which could be seen as falling prey to Goodhart's law \citep{Strathern1997}). 
These experiments offer evidence that g-SAEs learn more latents that have a more \textit{potent} and \textit{targeted} effect on their associated set of logits in arbitrary contexts. 

\setcounter{figure}{6}
\begin{figure*}[t]
    \centering
    \includegraphics[width=0.85\linewidth]{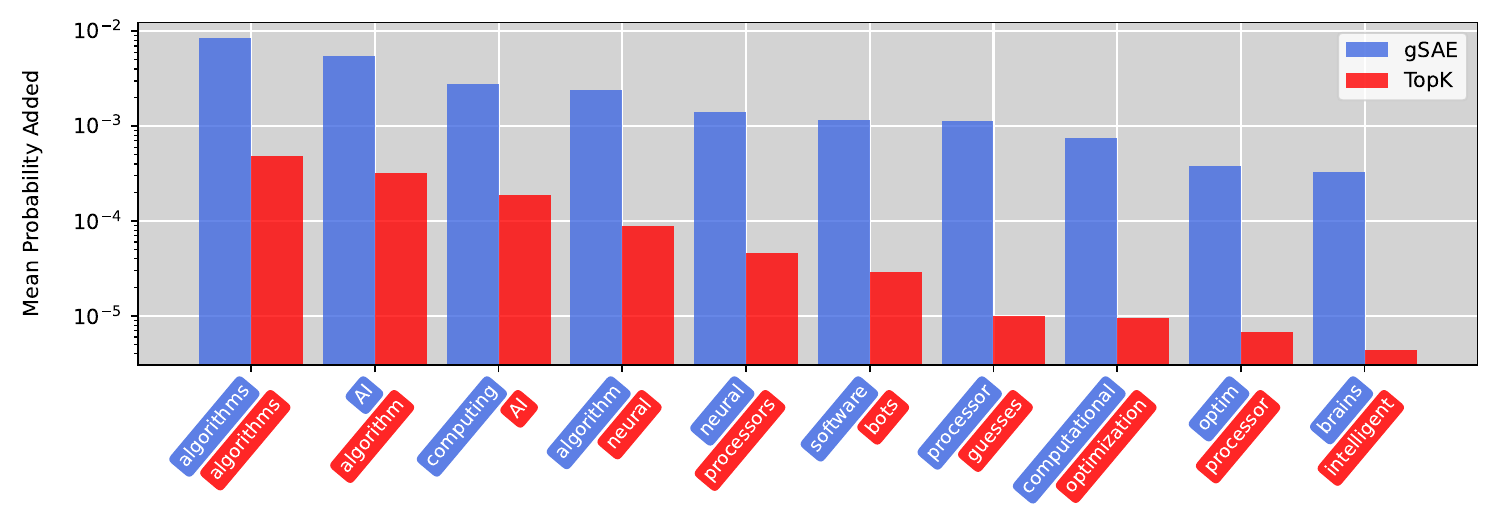}
    \caption{Average steering effects of the "AI" latents in each SAE. The mean amount added to all logits \textit{not} listed above was $7.3 \times 10^{-1}$ and $9.0 \times 10^{-1}$ for the g-SAE and TopK SAE, respectively. }
    \label{fig:concrete_comparison}
\end{figure*}

\setcounter{figure}{5}
\begin{figure}[H]
    \centering
    \includegraphics[width=0.95\linewidth]{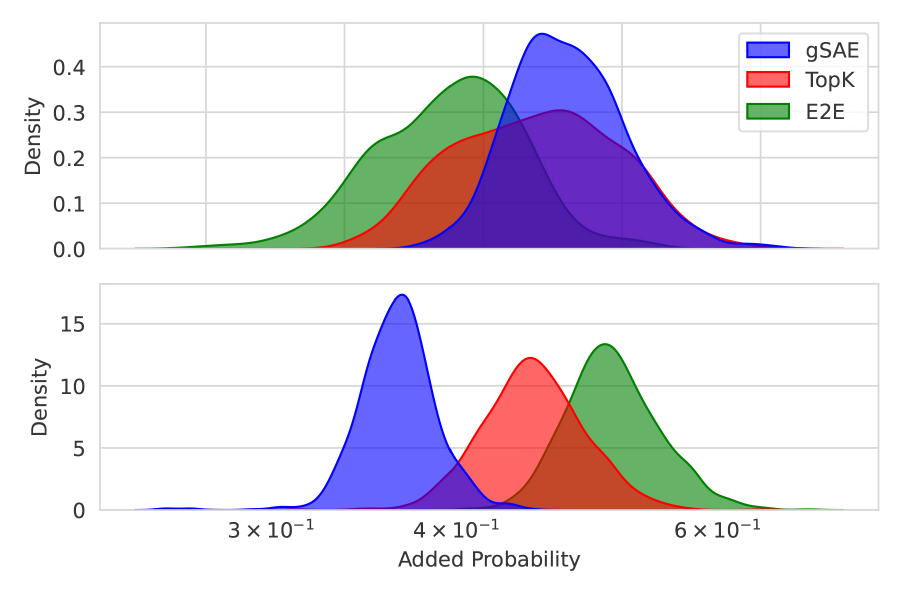}
    \caption{KDE plots of the added probability to associated logits for each SAE type due to the addition of steering vectors, where density is density of latents (Layer 7) \textbf{Top:} total probability added to associated logits. \textbf{Bottom:} total probability added to all but associated logits}
    \label{fig:KDEs}
\end{figure}

To demonstrate the steering capability of our g-SAEs with a specific example, we isolate the latent corresponding to the concept of “AI” in layer 5 for both g-SAE and TopK models. As detailed above, we form a steering vector by normalizing the selected latent to a fixed norm ($\alpha = 100$) and then we project this vector into the vocabulary space to identify its top 10 associated logits. Finally, we add the steering vector to the layer 5 activations for 100 different tokens in varied contexts and measure the average change in the probabilities of the associated logits. Our results shown in Fig. \ref{fig:concrete_comparison} confirm that g-SAEs steer model outputs toward AI-related concepts in a more targeted and potent manner than the standard TopK approach.

\subsection{Activity Density Analysis}

Intuitively, latents shouldn't be too specific, nor too general.
A sufficiently large SAE that learns a feature for every activation in its training data could achieve a very low reconstruction loss, however, the latents would be uninformative.
On the other hand, latents that activate for many (possibly unrelated) patterns of tokens are likewise uninformative. 
We analyze the activation density of a sample of latents from each type of SAE, which is a measure of how general or specific the latents of an SAE are.
Holding the sparsity of the SAEs constant at $L_0 = 32$, the cumulative distribution functions of latent activity (Fig. \ref{fig:latent_activity_analysis_combined}) show that g-SAEs tend to have fewer moderately high frequency latents which activate on more than $1\%$ percent of tokens, but more very high frequency latents which activate on more than $10\%$ of tokens.
We also see that the median latent of g-SAEs is more specific than than the others, with a median activation frequency of $0.24 \%$, compared to $0.43\%$ and $1.10\%$ for TopK and e2e SAEs, respectively. 

\subsection{Feature Splitting}

\begin{table}[H]
\centering
\begin{tabular}{lccc}
\hline
\textbf{SAE Type} & \textbf{Left Tail} & \textbf{Middle} & \textbf{Right Tail} \\
\hline
g-SAE & 2.070 & 2.538 & 2.861 \\
TopK & 2.187 & 2.313 & 2.366 \\
E2E & 2.221 & 2.502 & 3.428 \\
\hline
\end{tabular}
\caption{Average absolute directional derivatives ($\times 10^{-5}$) by position in latent similarity histogram for different SAE Types (Layer 7). Here we take a sample of 30 latents for each category and use 300 token sequences to compute the averages.}
\label{tab:directional_derivatives}
\end{table}

\setcounter{figure}{7}
\begin{figure*}[t]
    \centering
    \includegraphics[width=0.95\linewidth]{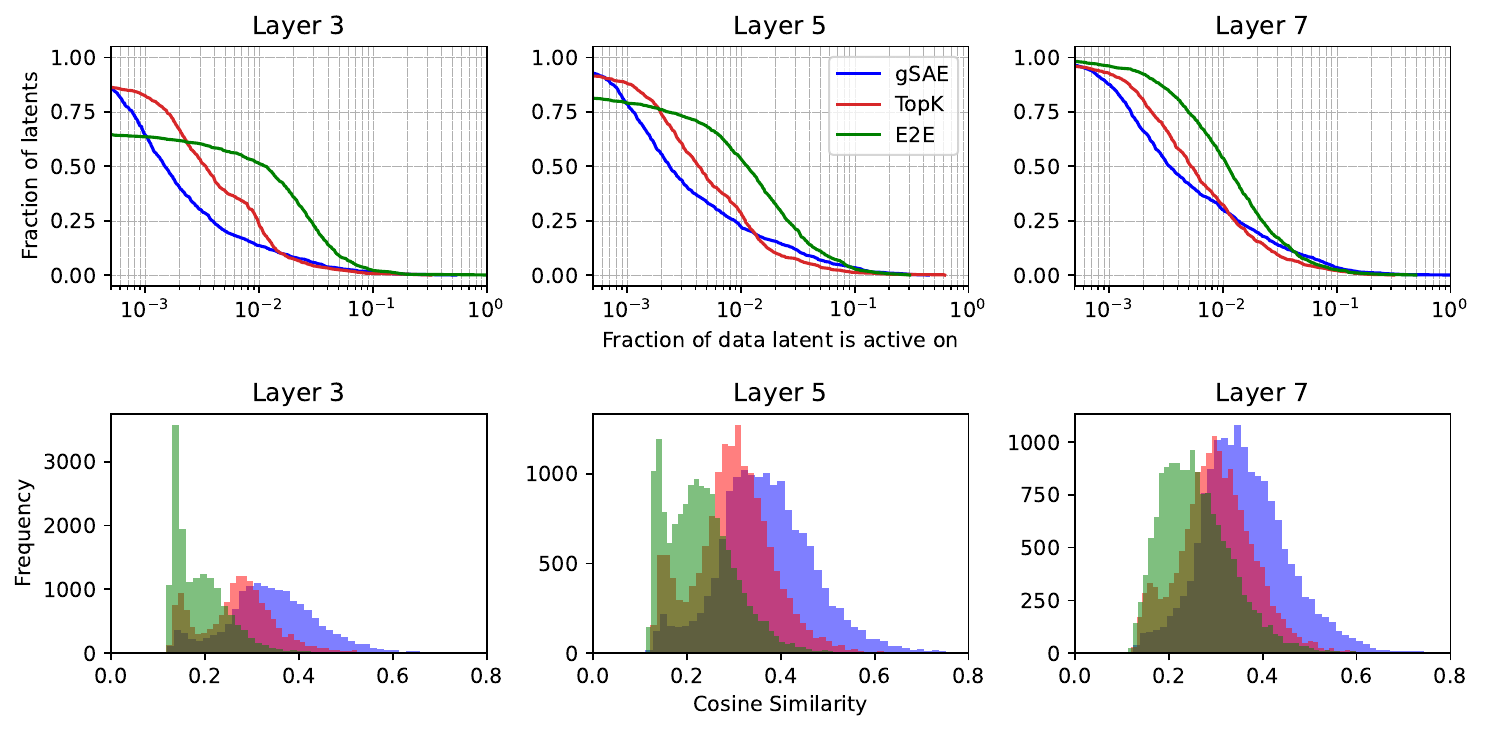}
    \caption{\textbf{Top:} Cumulative distribution functions of latent activity for the various SAEs at $L_0 = 32$ residual stream. \textbf{Bottom:} The cosine similarity of the most similar decoder column for every column in the SAE decoders. Note that the spikes at the left tail of the histograms likely correspond to dead latents. }
    \label{fig:latent_activity_analysis_combined}
\end{figure*}

\citet{Bricken2023Towards} and \citet{cunningham_sparse_2023} identified the phenomenon of \textit{feature splitting}, in which an SAE learns several latents which are associated with inputs that have little to no apparent semantic difference between them.
\citet{e2e_sae_braun} suggest that the cosine similarities between each SAE dictionary latent and next-closest latent in the same dictionary is a reasonable proxy for feature splitting, since latents which split features tend to be oriented in highly similar directions. 

In Fig. \ref{fig:latent_activity_analysis_combined} we see that the decoder column similarity histogram of g-SAEs has a higher mean than other SAE types.
To investigate this phenomenon further, we take a sample of latents from the left, middle, and right tails of the top decoder similarity histograms and calculate the average directional derivative in the direction of these decoder columns with respect to $\mathcal{L}(\mathbf{x})$ over many activations $\mathbf{x}$.
The results in Table $\ref{tab:directional_derivatives}$ suggest that in g-SAEs (as well as e2e SAEs), feature directions are more tightly clustered in high-influence directions.
This might indicate that the SAE is allocating more latents towards these directions, such that the density of latents in these high influence directions is higher.
Whether this phenomenon leads to more problematic instances of feature splitting in g-SAEs is unclear, and could be the subject of further research.

\subsection{Scaling with Model Size}

\begin{figure}[h]
    \centering
    \includegraphics[width=\linewidth]{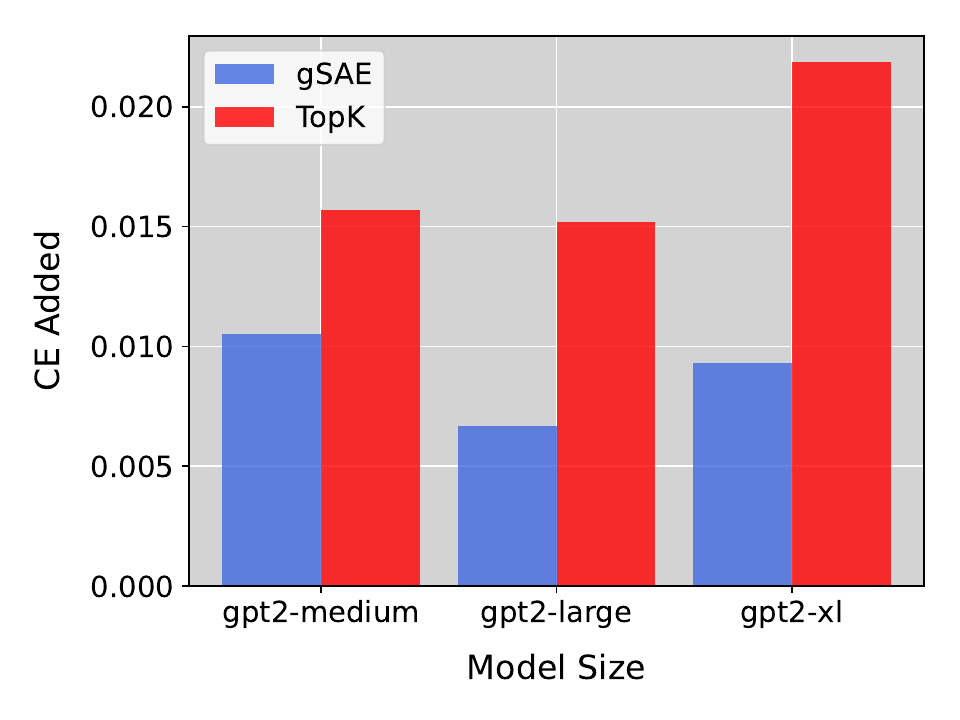}
    \caption{Average CE loss added ($\mathcal{L}_{\text{added}}$) by reconstructed activations from g-SAEs and TopK SAEs across various GPT-2 model sizes. \textit{Lower is better.} As model size increases, g-SAE reconstructions consistently yield lower CE loss added, indicating that they more faithfully capture causally relevant features.}

    \label{fig:gpt2_sizes}
\end{figure}

To assess the scalability of our approach, we trained both g-SAEs and TopK SAEs on different sizes of GPT-2. In each case, we extracted activations from a layer roughly three-quarters of the way through the model, and all other training hyperparameters—such as sparsity constraints, expansion factors, and learning rates—were held constant. We observe in Fig \ref{fig:gpt2_sizes} that on average, the reconstructed activations of g-SAEs are consistently associated with smaller changes in loss ($\mathcal{L}_{\text{added}}$) when propagated through the model, as compared to TopK SAEs. This finding confirms that the enhanced encoding of causally relevant information afforded by g-SAEs persists across model scales.

\subsection{Manual Interpretability}

To ensure that the increased functional influence of g-SAE latents don't come at the cost of their monosemanticity as representations, we follow a double blind experimental setup similar to that of 
\citet{rajamanoharan2024improvingdictionarylearninggated} and \citet{rajamanoharan2024jumpingaheadimprovingreconstruction} to measure the interpretability of g-SAE latents according to human raters.
We selected three SAE architectures (TopK, E2E TopK, g-SAE), trained on the residual stream of GPT-2 small at multiple layers (3, 5, 7, 9), for a total of 16 SAEs. We held TopK sparsity at 32, and our input expansion factor at 20.

Our 3 human raters were presented with samples of activating examples in a random order, with summary information for each feature, and tasked with deciding whether each feature is monosemantic, with answers being 'Yes', 'Maybe' and 'No'. We collected 384 samples, or 32 per SAE.

\begin{figure}[h]
    \centering
    \includegraphics[width=0.95\linewidth]{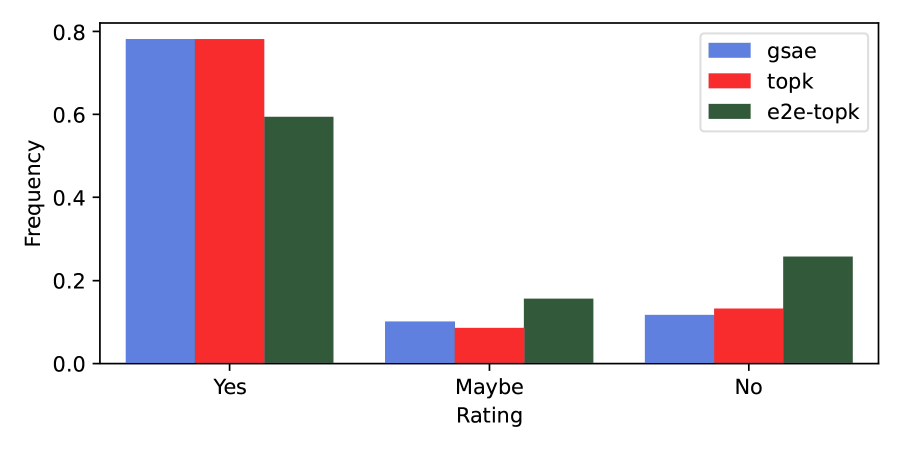}
    \caption{Human evaluation scores for feature interpretability.}
    \label{fig:manual_evaluation}
\end{figure}

\textbf{Results}\ \ We present our results in Fig. \ref{fig:manual_evaluation}. Our reviewers rated g-SAEs and TopK SAEs as similarly interpretable, while e2e-TopK SAEs were slightly less so. 

\section{Conclusion}
Our experiments show that introducing gradient information into the activation function of SAEs produces improvements in output explained at a given expansion size and sparsity. 
Additionally, the resulting SAE latents appear to have an increased ability to steer model outputs in a concentrated manner in arbitrary contexts by consistently increasing the probability of the associated set of tokens. 
We believe this improvement is significant both for practical applications aiming to exert more fine-grained control over model outputs, and for interpretability applications that aim to uncover the primary units of models' representations and actions. 
We find that this improvement comes at no measurable cost to semantic interpretability to human reviewers.

We believe that these developments will lead to dictionary learning that more fully accounts for both the representational and causal aspects of models' features, ultimately contributing to a better understanding of the fundamental workings of LLMs.

\section{Limitations}
Here we discuss limitations of g-SAEs, along with a discussion of the ethical considerations.

\begin{itemize}
\item \textbf{Activation Plateaus:} There is evidence that models may be robust to small perturbations around activations, as a consequence of emergent robustness against small noise \citep{giglemiani2024evaluatingsyntheticactivationscomposed}.
Despite our evidence that gradients in feature directions are higher than those of baseline directions, it may be that such regions of noise-robustness are locally flat enough that the local gradient doesn't provide information about the sensitivity of the model to larger perturbations in the vicinity.
If this is the case, we would plausibly want to add a term to the activation function that privileges directions with a large higher order derivatives, but it isn't clear how to do so efficiently. 
\item \textbf{Computational Complexity:} One of the potential drawbacks of g-SAEs is their theoretical computational cost, as both training and running inference on them requires a partial forward pass and backpropagation.

In practice, we found that this was not a computational bottleneck and there wasn't a significant change in cost compared to TopK SAEs when training on the different sizes of GPT-2.
\item \textbf{Evaluations:} There is a significant need within the community for more thorough evaluations that truly test the quality of SAEs and their potential effect on model behavior. Recent work is making progress in this direction. \citet{karvonen_measuring_2024} propose new quantifiable measures for monosemanticity and feature extraction quality using a toy model. \citet{makelov_sparse_2024} demonstrate the use of SAEs for model control in the Indirect Object Identification (IOI) task, providing a task to test the quality of SAEs on. While these studies provide valuable insights, further research is needed to develop comprehensive evaluation frameworks that can be applied more generally to SAEs and their quality. In the context of g-SAEs, it remains to be studied how much increased steerability on the \textit{token} level translates to increased steerability on higher levels of abstraction. 
\end{itemize}
\subsection{Ethics}
The ethical considerations of this paper reflect the broader ethical considerations around LLMs.
We use publicly available datasets and models which can represent potentially problematic biases. 
However, our work does not contribute in any specific way to the propagation of potentially negative ideas present in the data.

A potential negative use case of g-SAEs, as with other sparse autoencoders, is to steer advanced models towards producing harmful outputs.
However, we believe that while progress in interpretability likely inherently comes with the possibility of more fine grained control over AI systems for both good and ill, the upside to a deeper understanding of deep learning systems is currently net positive. 

\section{Acknowledgements}

We thank Thomas Dooms and Daniel Wilhelm for their smol-sae library which we modified and used for the training of our SAEs. We also thank Callum McDougal for the sae\textunderscore vis library which was used in our manual interpretability experiments.

\bibliography{custom}

\end{document}